# Imaging the time series of one single referenced EEG electrode for Epileptic Seizures Risk Analysis

Tiago Leal, António Dourado (*Senior Member, IEEE*), Fábio Lopes, César Teixeira

*Abstract*— The time series captured by a single scalp electrode (plus the reference electrode) of refractory epileptic patients is used to forecast seizures susceptibility. The time series is preprocessed, segmented, and each segment transformed into an image, using three different known methods: Recurrence Plot, Gramian Angular Field, Markov Transition Field. The likelihood of the occurrence of a seizure in a future predefined time window is computed by averaging the output of the softmax layer of a CNN, differently from the usual consideration of the output of the classification layer. By thresholding this likelihood, seizure forecasting has better performance. Interestingly, for almost every patient, the best threshold was different from 50%. The results show that this technique can predict with good results for some seizures and patients. However, more tests, namely more patients and more seizures, are needed to better understand the real potential of this technique.

## I. Introduction

Epilepsy is one of the most prevalent neurological diseases, affecting approximately 1% of the world's population, of all regions, ages and conditions. About one third of these suffer from refractory epilepsy, for which surgery or antiepileptic drugs are ineffective, and must live with seizures that can happen anytime, anywhere. Their lives are very constrained, and full social integration is difficult. Some decades of research have not yet discovered an algorithm and a transportable device implementation to warning them of a coming seizure.

The brain state is usually measured by the Electroencephalograms (EEG) that are acquired by a set of electrodes implanted inside the brain (iEEG) or clued to the scalp (sEEG), the preferred solution since it is noninvasive (actually there are some developments to insert the electrodes between the scalp and the brain bone, to improve the signal/noise ration). When using machine learning models as classifiers, the brain state must belong to a certain class or pattern. The most common are the interictal, the preictal, the ictal (the seizure state), and postictal states. In this work only the interictal and the preictal are considered because it aims to forecast the seizures, and not to detect them. The pre-ictal is the period before the seizure, and the interictal is the period where there is no seizure nor in the short future neither in the short past. In this work scalp EEG signals are used. These signals are nonstationary, very sensitive to noise, artifacts, and other interferences. Unfortunately, the seizure generation processes are different from patient to patient, and even from seizure to seizure for the same patient, which is an additional challenge to develop seizure forecasting algorithms, that must be patient specific. This helps to explain why, after 30 years of research, no one found an algorithm with acceptable performance in real life.

Most works about seizure forecasting actually use machine learning models and the most relevant performance measures are the Sensitivity and False Positive Rate per hour (FPR/h), and these are used in this work.

The reviews [1] and [2] give a good general overview of past research. In this work deep learning is used without any previously extracted features. There are a few studies using also deep learning [3][4], some reporting good results in testing in limited collected data. But the lack of access to a vast and continuous long-term EEG data (big data), caused the results not to be reproducible in real life.

Research supported by FCT Foundation.

A. Dourado (correspondent author) and C. Teixeira are with the Department of Informatics Engineering, University of Coimbra, and with CISUC, 3030-290 Coimbra Portugal, phone + 351 239 79 0000, e-mail dourado@dei.uc.pt, cteixei@dei.uc.pt. T. Leal (leal@student.dei.uc.pt) and F. Lopes (fabioacl96@gmail.com) are with CISUC.

Using multi-channel EEG, transforming the multidimensional time-series into images and using deep learning has recently been used for seizures detection [16] and seizure risk assessment [17].

In our previous work [17] time series from 19 electrodes were used. However, a lower number of electrodes is more comfortable for the patient, to have the EEG continuously being read and give to an algorithm running in a miniaturized transportable device.

In this work we test the fewest possible number of electrodes - just one plus the reference electrode (the original reference is the AF7 in the standard 20x20 montage)– and train a Convolution Neural Network adapted from the AlexNet [12] with images built from the captured unidimensional time series.

The EEG time-series of one single scalp electrode, the one closest to the epileptic focus, are segmented by seconds, and each second is transformed into an image, which will then be used to train a Convolutional Neural Networks (CNN) to forecast seizures, having two classes in the output: the interictal and the preictal.

Three techniques are used to transform the signals into images: (i) Recurrence Plot [5], (ii) Gramian Angular Field [6], and (iii) Markov Transition Field [6].

The output of the softmax layer returns the probability of the image given to the CNN being in the interictal or in the preictal class. Likelihood thresholds are created over the mean of softmax layer outputs for the last minute of data (60 images) to measure the likelihood of an oncoming seizure. The best threshold for each patient will then be found by trial-and-error, by checking its ability to forecast seizures.

This work is the continuation of our previous work [17] where we used multidimensional time series with 19 electrodes.

## II. METHODOLOGY

16 patients from the European Epilepsy Database [7] were used. All of them were pre-processed by the method explained in section II of Lopes et al. [8]. The EEG signals were filtered using a 0.5-100 Hz bandpass 4th-order Butterworth filter and a 50 Hz 2nd-order notch filter, to remove noise and power line interference. The divided EEG segments of the signal were then average referenced with extended-infomax Independent Component Analysis (ICA) from EEGLAB [15]. The resulting independent components (ICs) were visually inspected by experts to eliminate noisy components. After that inspection, the signal was reconstructed with the non-noisy ICs.

Patients' seizures were divided into training and testing; around two-thirds of the seizures were used for normal training of the CNN as a classifier. Testing was not made in the usual way. Instead, testing seizures were used to find the best likelihood threshold by trial and error.

### A. Training

Firstly, the single electrode to be used was chosen from the data in Teixeira et al. [9]. For each patient, the electrode closest to the seizure focus was chosen, resulting in Table 1. Before transforming the electrode signal into images, normalization was applied. The whole signal was divided by the maximum absolute value in the time series, reducing it to the range [-1, 1], that is the input needed for the three methods used.

*Table I – Electrodes selected for each patient.*

| Patient Number | Electrode closest to more seizures focus | Number of seizures |
|---|---|---|
| 8902 | F7 | 5 |
| 11002 | F8 | 4 |
| 26102 | C3 | 4 |
| 32702 | T7 | 5 |
| 45402 | T8 | 4 |
| 52302 | F7 | 4 |
| 53402 | F4 | 4 |
| 55202 | F8 | 8 |
| 64702 | T8 | 5 |
| 94402 | F7 | 7 |
| 95202 | T7 | 7 |
| 101702 | F8 | 5 |
| 112802 | T8 | 6 |
| 113902 | F8 | 6 |
| 114902 | P8 | 7 |
| 123902 | T7 | 5 |

The interictal parts of the training signal were segmented by a sliding window of 256 samples, corresponding to one second, since the sampling frequency is 256 Hz. The preictal interval was segmented also by a sliding window of the same size, but with 50% overlapping, to create more preictal images. Class balancing was done by selecting the same amount of preictal segments as interictal ones, so that in training both classes had the same number of images. The algorithm tries to select the same number of interictal segments from before each seizure, to increase the generalization capability of the CNN, preventing a bias towards a particular seizure. For each selected segment, the three methods applied to create images are briefly described in the following.

(i) *Recurrence Plot* [5] describes natural time correlation between points in different instants. It is built by extracting trajectories in an embeding dimension of the time series, and the binarized pairwise distance between those trajectories. For more details see [5]. It was implemented using the RecurrencePlot function from the pyts module for Python [10], with all parameters in default.

Figure 1 shows two segments of the single chanel EEG with the duration of one second, one interictal, the other ictal, normalized to [0 1].

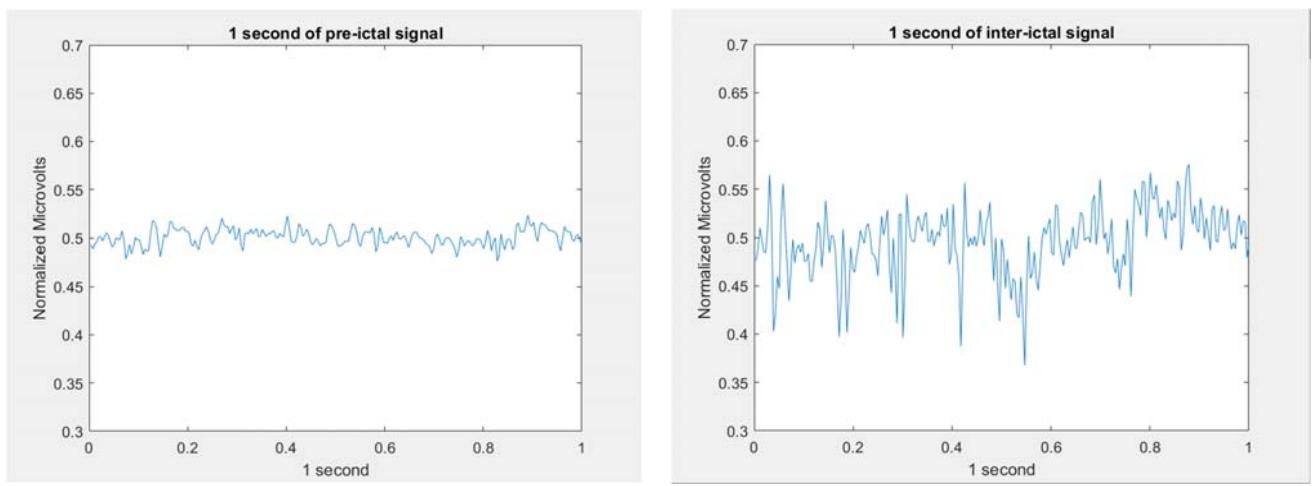

*(a)* *(b)*

*Figure 1- Example of 1s segments; (a) preictal, (b) interictal.*

Figure 2 shows the figures produced by the Recurrence Plot with the segments of Figure 1.

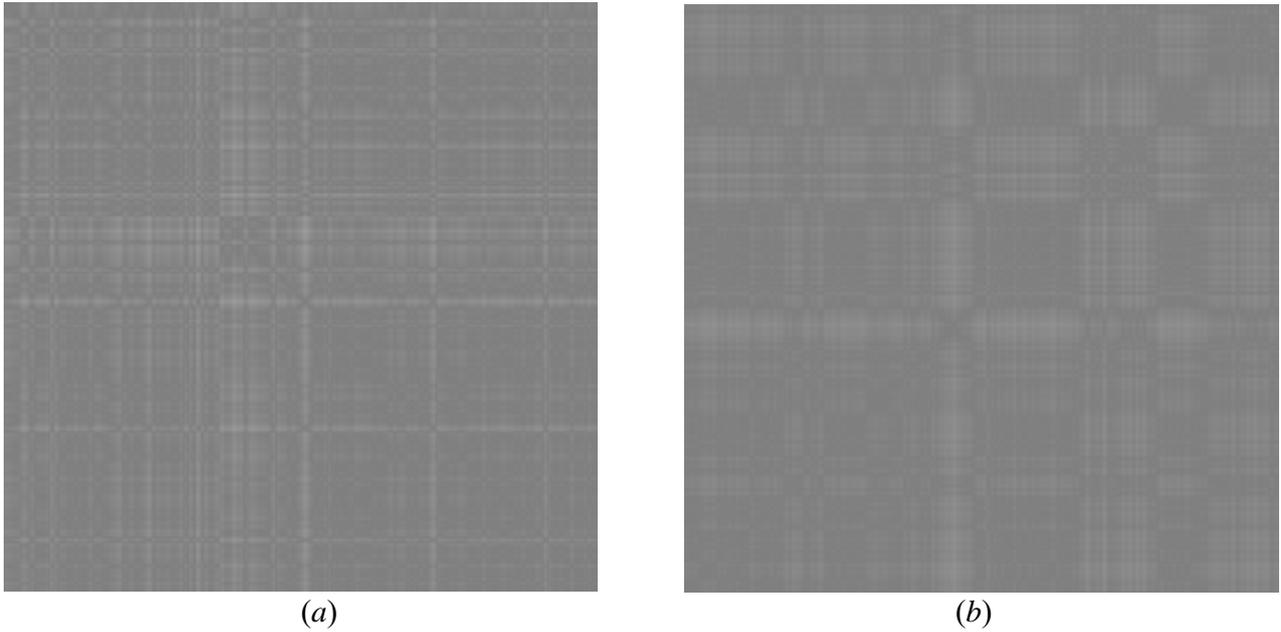

*Figure 2. Images created by the Recurrence Plot: (a) preictal, (b) interictal*

*(ii) Gramian Angular Field* [6] is based in the polar coordinates of the time series (instead of rectangular coordinates). The time series is previously rescaled to the interval [a, b] with $-1 \leq a < b \leq 1$. With the polar coordinates a Gramian matrix is computed that can be interpreted as the pixels of an image. For more details see [6]. The Summation variant was used in this work. It was implemented using the GramianAngularField function from pyts [10], with all parameters in default. Figure 3 shows the images corresponding to to the time series of Figure 1.

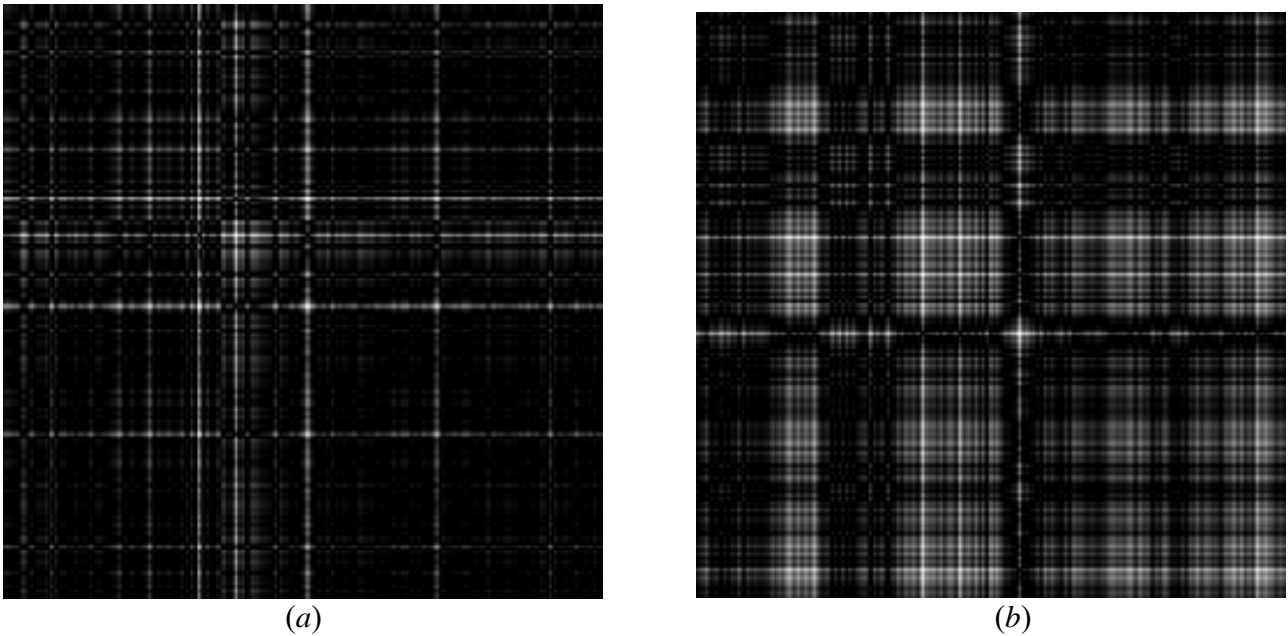

*Figure 3 – Images from Gramian Angular Field; (a) preictal, (b) interictal.*

*(iii)Markov Transition Field* divides the time series into a number of quantile bins, assigns each point to a bin and after some operations computes a Markov transition matrix followed by a Markov Transition Field matrix that can be interpreted as one image. For a detailed development see [6]. The MarkovTransitionField function from pyts [10], with all parameters in default, was used to implement it. Figure 4 results from its aplication to the time series of Figure 1.

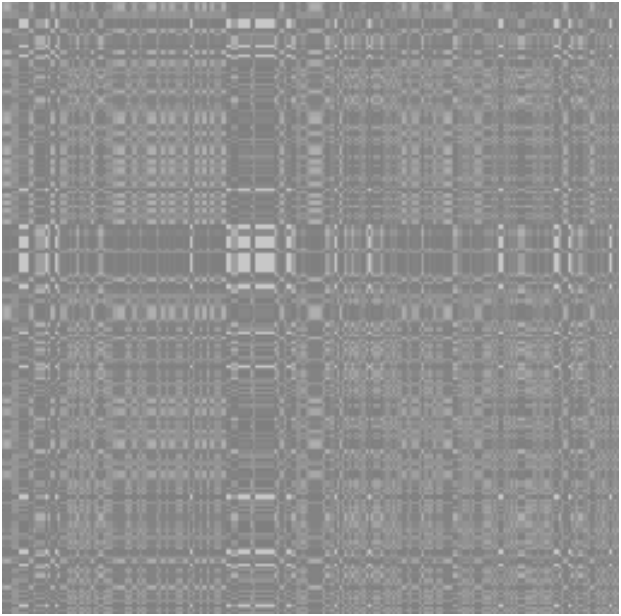
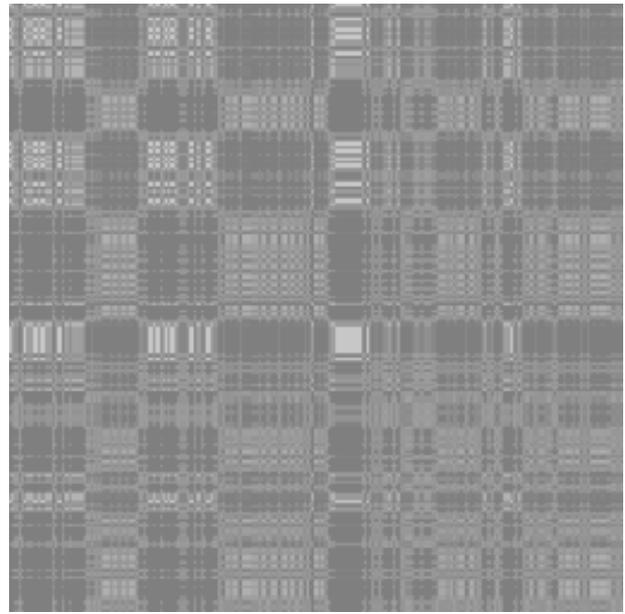

(*a*)                                                                                      (*b*)

*Figure 4 – Images from Markov Transition Field; (a) preictal, (b) interictal.*

Because of the methods used to save the images in the disk (IPL's convert() with 'RGB' parameter), the images were saved having dimensions of 256x256x3.

With the training images dataset ready, they were used to train a CNN. An adaptation of the well-known AlexNet [11] was used. It contains 25 layers and in its original form needs 227x227x3 image inputs. This work used the MATLAB Deep Learning Toolbox alexnet method [12] to apply it. The original network can distinguish between 1000 classes of objects. In this work, however, it was adapted to distinguish between only two classes, and find if an image belongs or not to the preictal class. Also, as the images in this work are 256x256x3, the alexnet input layer was changed to accept such dimensions. The $23^{rd}$ layer was changed to a Fully connected layer with 2 classes (two outputs), the 24th to a softmax layer with two inputs and two outputs, and the 25th into a classification layer with two outputs. The softmax layer will be the one used to calculate the likelihood, while classification layer is used as an aid to train the network.

A mini-batch size of 64 and a learning rate of 0.001 were chosen, based on Matlab suggestions, and 50 epochs were used in training; however, the training curve usually stabilized before 50 epochs. Since the clinical knowledge does not allow yet to fix the preictal time, four preictal times of 10, 20, 30, and 40 minutes were empirically considered. So, for each patient, 12 networks were trained and tested, by combining the 3 types of images with the 4 preictal times.

## B. Finding the best likelihood thresholds

For the seizures in the second group (the testing group), the whole electrode signal was segmented into segments of 1s (1x256 time series), with no overlapping. After the segmentation, images of the three types were created from each segment and labelled according to the class they belong to.

Each image was given to the CNN of the respective type and preictal time, and instead of the classification layer output, the softmax layer preictal output was considered as a seizure likelihood occurrence indicator. This layer returns 2 values: the probability of the image belonging to the interictal class, and the same for the preictal class. As one is the complement of the other, only the second value was considered.

The left side of Figure 5 covers 60 s of one signal, i.e., the preictal output of the softmax layer for each image, and the black line the average of the last 60 images. The black line can be considered as the likelihood at each instant of a coming seizure. The right side of Figure 4 shows the preictal softmax output for the interictal (blue) and preictal (green) related to one seizure; this output varies considerably even for consecutive images. To overcome the difficulty that this causes, the mean of the last minute of probabilities was considered. At one moment the likelihood of an incoming seizure is measured by the mean of the probabilities obtained for the last 60 images (the other 59 being the ones before the current). This value can be seen in Figure 5 by the black line.

The green zone corresponds to the real preictal, and the blue zone to the real interictal. Note that this is testing labeled data. The likelihood (the black line) in Figure 4 b) is higher in most instants of the preictal time, as it was expected.

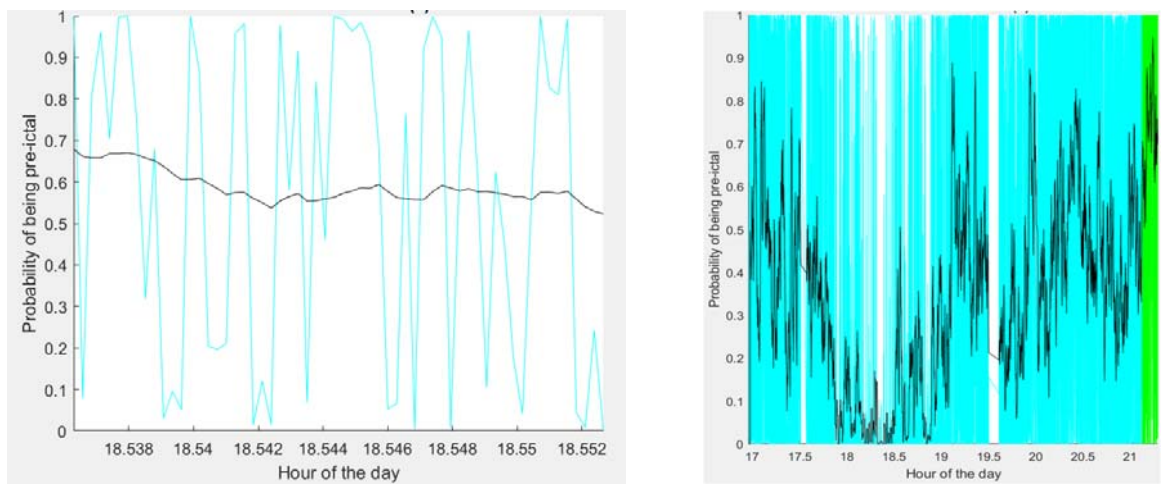

*Figure 5 - Likelihoods given by preictal softmax layer output to each image being preictal and the mean of the last minute (the black line) for a period of 60 seconds (left) and for a whole seizure (right).*

That likelihood was then used to predict incoming seizures. A threshold $Z$ was created so that if the likelihood is over it, the current instance is considered of the preictal class. Then, to predict, Firing Power technique was applied [14]. Only if more than $Y$ % of the images in the last $X$ minutes ($X$ being the preictal time) were considered preictal, a seizure alarm would be raised. When raised, the Seizure Prediction Horizon (SPH) and the Seizure Occurrence Period (SOP) would follow [14]. The SPH was always 5 minutes, while the SOP was half the considered preictal time. The Firing Power decreases the number of false alarms. The $Y$ % works like a threshold. So, there are two thresholds: the likelihood threshold $Z$ and the prediction threshold $Y$.

To summarize, images are given one by one to the trained network of the respective size and preictal time. The probability that the softmax layer of the network returns for that image to be preictal is obtained. Together with the same probabilities of the last 59 images, the mean of the probabilities is computed. That mean is the likelihood that a seizure could occur soon. To predict, a threshold is needed such that, if in the last *X* minutes, *Y* % of the instances (means of the last minutes) are considered above it, an alarm is fired.

III. EXPERIMENTS

The objective of the experiments was to find, for each patient, which seizure occurrence likelihood threshold had the best performance in predicting seizures. Several values for the best likelihood threshold *Z* and prediction threshold *Y* were tested. For *Z*, values from 0.05 to 0.9, with a step of 0.05 (0.05, 0.1, 0.15, …, 0.9) were tested. For *Y*, values from 0.2 to 0.9 were considered with a step of 0.1. Each of these were tested for different preictal times and for each of the image types used to find the best results, in a trial-and-error approach. In Tables II, III, and IV the combination of these variables that performed the best in predicting can be seen for each image type.

The sensitivity SE is related to the well predicted seizures, it is frequently used in this type of studies. It is also known as Recall, or True Positive Rate and given by (1)

$$SE = \frac{TP}{TP + FN} = \frac{TP}{P} \qquad (1)$$

where *TP* is *True Positive*, *FN False Negative*, and *P* all the labeled positives.

*False Positive Rate FPR/h* is the number of false alarms per hour, on average, for the test set when testing and for the training set when training. Note that the following results are for the testing set of each patient.

*Table II – Best thresholds with Recurrence Plot transformation.*

| Patient | Pre-ictal minutes | Prob. Thresh. Z | FP Thresh. Y | Sens. SE | FPR/h |
|---|---|---|---|---|---|
| 8902 | 40 | 0,85 | 0,8 | 1 | 0,072 |
| 11002 | 40 | 0,1 | 0,9 | 1 | 0,296 |
| 26102 | 40 | 0,2 | 0,3 | 1 | 0,257 |
| 32702 | 10 | 0,6 | 0,4 | 1 | 0,099 |
| 45402 | 20 | 0,7 | 0,7 | 1 | 0,325 |
| 52302 | 10 | 0,35 | 0,7 | 1 | 0,071 |
| 53402 | 20 | 0,05 | 0,65 | 1 | 0,282 |
| 55202 | 40 | 0,55 | 0,25 | 1 | 0,289 |
| 64702 | 40 | 0,1 | 0,8 | 1 | 0,289 |
| 94402 | 40 | 0,7 | 0,2 | 0,667 | 0,289 |
| 95202 | 20 | 0,65 | 0,3 | 1 | 0,050 |
| 101702 | 30 | 0,25 | 0,85 | 1 | 0,258 |
| 112802 | 40 | 0,5 | 0,8 | 1 | 0,351 |
| 113902 | 10 | 0,2 | 0,75 | 1 | 0,381 |
| 114902 | 40 | 0,5 | 0,25 | 1 | 0,246 |
| 123902 | 20 | 0,3 | 0,9 | 1 | 0,141 |

*Table III - Best thresholds with Gramian Angular Field transformation.*

| Patient | Pre-ictal minutes | Prob. Thresh. Z | FP Thresh. Y | Sens. SE | FPR/h |
|---|---|---|---|---|---|
| 8902 | 40 | 0,9 | 0,4 | 1 | 0,064 |
| 11002 | 20 | 0,3 | 0,35 | 1 | 0,209 |
| 26102 | 40 | 0,35 | 0,25 | 1 | 0,251 |
| 32702 | 10 | 0,6 | 0,6 | 1 | 0,114 |
| 45402 | 20 | 0,75 | 0,8 | 1 | 0,366 |
| 52302 | 10 | 0,45 | 0,75 | 1 | 0,308 |
| 53402 | 20 | 0,1 | 0,5 | 1 | 0,275 |
| 55202 | 40 | 0,45 | 0,45 | 1 | 0,289 |
| 64702 | 40 | 0,1 | 0,85 | 1 | 0,293 |
| 94402 | 40 | 0,05 | 0,2 | 1 | 0,419 |
| 95202 | 20 | 0,5 | 0,2 | 1 | 0,081 |
| 101702 | 40 | 0,6 | 0,2 | 1 | 0,283 |
| 112802 | 40 | 0,5 | 0,55 | 1 | 0,376 |
| 113902 | 40 | 0,35 | 0,4 | 1 | 0,420 |
| 114902 | 30 | 0,25 | 0,6 | 1 | 0,345 |
| 123902 | 40 | 0,5 | 0,2 | 1 | 0,266 |

*Table IV - Best thresholds with Markov Transition field transformation.*

| Patient | Pre-ictal minutes | Prob. Thresh Z. | FP Thresh Y | Sens. SE | FPR/h |
|---|---|---|---|---|---|
| 8902 | 40 | 0,85 | 0,8 | 1 | 0,065 |
| 11002 | 20 | 0,3 | 0,25 | 1 | 0,290 |
| 26102 | 40 | 0,25 | 0,25 | 1 | 0,251 |
| 32702 | 10 | 0,6 | 0,25 | 1 | 0,137 |
| 45402 | 20 | 0,55 | 0,9 | 1 | 0,389 |
| 52302 | 20 | 0,45 | 0,4 | 1 | 0,396 |
| 53402 | 30 | 0,1 | 0,8 | 1 | 0,223 |
| 55202 | 40 | 0,55 | 0,45 | 1 | 0,289 |
| 64702 | 20 | 0,4 | 0,8 | 1 | 0,219 |
| 94402 | 40 | 0,05 | 0,25 | 1 | 0,420 |
| 95202 | 10 | 0,5 | 0,2 | 1 | 0,088 |
| 101702 | 40 | 0,5 | 0,6 | 1 | 0,342 |
| 112802 | 40 | 0,45 | 0,75 | 1 | 0,354 |
| 113902 | 40 | 0,4 | 0,25 | 1 | 0,415 |
| 114902 | 20 | 0,3 | 0,9 | 1 | 0,355 |
| 123902 | 40 | 0,3 | 0,35 | 1 | 0,395 |

Patients 55202, 94402, 95202, 112802, 113902, and 114902 had around 13.5 hours (three seizures) of testing data while the rest had around nine hours (two seizures). In the header, FP means Firing Power.

## IV. Discussion

For Recurrence Plot transformation, 100% sensitivity (SE=1) was obtained for every patient but one. Among those, 12 have FPR/h below 0.3 and 5 below 0.15, the maximum recommended by Winterhalder et al. [13]. As for Gramian Angular Field and Markov Transition Field, these achieved 100% sensitivity in every patient. However, overall, FPR/h was higher than in Recurrence Plot. Both predict 100% seizures with an FPR/h below 0.15 for three patients, and 10 times below 0.3. For the patient where 100% isn't achieved in Recurrence Plot, these two methods only achieve so with a very high FPR/h.

All optimal parameters vary a lot from patient to patient, which is something already known in seizure prediction community. Preictal times chosen are 40 minutes for most of the 16 patients, which means that there is a possibility that if experiments were done with higher values, results for these patients eventually could be further improved.

As for the likelihood thresholds, for some patients, the 50% used by the classification layer of the CNN is indeed the optimal choice. Out of these, only in 1 (95202, with Markov and Gramian) this threshold predicts 100% of the seizures with FPR/h below 0.15. The fact that several of the thresholds deviate from the 50%, means that analyzing the output of the softmax layer, and considering a likelihood calculated based on it, could help improve CNN's performance in prediction in a number of patients by adjusting the likelihood threshold. Some patients are more propense to suffering a seizure when the likelihood is lower, while in others, the opposite happens. In some cases, present in the table, the results deviate considerably from 0.5. This may happen because the seizures they were trained with may be too different from the ones used to find the thresholds. It highlights the need to more experiments by, for example, doing more runs of this experiments changing the roles (training or testing) that the seizures take each time.

## V. Conclusion and future work

Generally, the best likelihood thresholds found can be used to predict and help in the cases where the default 50% thresholds may not be the most correct. The fact that they were able to do it with acceptable performance in a considerable number of patients and with only one electrode is even more noteworthy and magnifies the potential this method has. If a seizure prediction algorithm can predict using only one electrode (referenced), quality of life of the patient can be improved, as it would improve his comfort. By calculating the threshold from which the patient is more propense to suffering also helps improving interpretability, as CNN is a black-box classifier.

However, more tests and more data are still needed, and it is important to overcome the problem caused by the number of seizures needed to find a perfectly adjusted threshold for a patient. Doing more runs with core data of this experiment, using other methods to calculate the likelihood or other methods to transform the signal into images, or even other CNN structures are all paths that can be followed to improve these results.

## Acknowledgment

Tiago Leal acknowledges the 8 months research grant from CISUC. This work is funded by the FCT - Foundation for Science and Technology, I.P./MCTES, through national funds (PIDDAC), within the scope of CISUC R&D Unit - UIDB/00326/2020 or project code UIDP/00326/2020.